%% file: main.tex
\documentclass[preprint,1p,times]{elsarticle}

\usepackage{amssymb, amsthm, makeidx, enumerate}
\usepackage{amsmath, latexsym, amscd, amsfonts}
\usepackage{times} 
\usepackage{textcomp}
\usepackage{wasysym}

\usepackage{changepage}

\usepackage{listings} 
\makeindex
\usepackage{pgf, tikz, mathrsfs, pgfplots, tkz-tab}
\usetikzlibrary{calc, angles, quotes, intersections, positioning, patterns, arrows}

\usepackage{color, fancyhdr, graphicx, wrapfig}
\usepackage[unicode]{hyperref}
\usepackage{lscape}

\usepackage{caption}
\usepackage{subcaption}

\usepackage{algorithm}
\usepackage{algpseudocode}

\theoremstyle{definition}
\theoremstyle{definition}
\theoremstyle{definition}
\newcounter{danhsovidu}[section] 

\usepackage{subcaption}
\newcommand{\tron}[1]{\ensuremath{\left( #1 \right)}}
\newcommand{\vuong}[1]{\ensuremath{\left[ #1 \right]}}

\newcommand{\chuan}[1]{\ensuremath{\left\| #1 \right\|}}

\usepackage{graphicx}
\usepackage{pgf, tikz}
\usepackage{mathrsfs}
\usetikzlibrary{arrows}
\usetikzlibrary{calc}
\usepackage{eso-pic}
\usepackage{float}
\usepackage{booktabs}

\usetikzlibrary{arrows.meta,shapes.arrows,shapes.geometric,arrows,shapes,matrix,positioning}

\tikzstyle{sa} = [rectangle, rounded corners, 
minimum width=0.9cm, 
minimum height=0.9cm,
text centered, 
draw=black, 
fill=cyan!40]

\tikzstyle{lo} = [rectangle, rounded corners, 
minimum width=2cm, 
minimum height=1cm,
text centered, 
draw=black, 
fill=lime!60]

\tikzstyle{to} = [very thick,->,>=stealth]

\tikzset{ 
table1/.style={
  matrix of nodes,
  row sep=-\pgflinewidth,
  column sep=-\pgflinewidth,
  nodes={rectangle,draw=black,text width=3ex,align=center},
  text depth=0.3ex,
  text height=1.5ex,
  nodes in empty cells,ampersand replacement=\&
  },
texto/.style={font=\small\sffamily},
title/.style={font=\small\sffamily}
}

\tikzset{ 
table2/.style={
  matrix of nodes,
  row sep=-\pgflinewidth,
  column sep=-\pgflinewidth,
  nodes={rectangle,draw=black,text width=4ex,align=center},
  text depth=0.3ex,
  text height=2ex,
  nodes in empty cells,ampersand replacement=\&
  },
texto/.style={font=\small\sffamily},
title/.style={font=\small\sffamily}
}

\begin{document}

\begin{frontmatter}



\title{Missing Data Imputation using Neural Cellular Automata}

\author{Trung-Tin Luu}
\author[1]{Thanh-Binh Nguyen}
\author[2]{Minh-Man Ngo}

\affiliation[1]{
    organization={University of Science},
    city={Ho Chi Minh City},
    country={Vietnam}
}
\affiliation[2]{
    organization={University of Banking},
    city={Ho Chi Minh City},
    country={Vietnam}
}

\input{content/0.abstract}

\end{frontmatter}




\input{content/1.introduction}

\input{content/2.review}

\input{content/3.preliminary}

\input{content/4.methodology}

\input{content/5.experiments}

\input{content/6.conclusion}





\bibliographystyle{elsarticle-harv}
\bibliography{reference}

\end{document}

%% file: content/0.abstract.tex
\begin{abstract}
When working with tabular data, missingness is always one of the most painful problems. Throughout many years, researchers have continuously explored better and better ways to impute missing data. Recently, with the rapid development evolution in machine learning and deep learning, there is a new trend of leveraging generative models to solve the imputation task. While the imputing version of famous models such as Variational Autoencoders or Generative Adversarial Networks were investigated, prior work has overlooked Neural Cellular Automata (NCA), a powerful computational model. In this paper, we propose a novel imputation method that is inspired by NCA. We show that, with some appropriate adaptations, an NCA-based model is able to address the missing data imputation problem. We also provide several experiments to evidence that our model outperforms state-of-the-art methods in terms of imputation error and post-imputation performance.
\end{abstract}




%% file: content/1.introduction.tex
\section{Introduction}
\label{introduction}

There is no doubt that data plays a crucial role in this modern world. In numerous business and scientific applications, data is the foundation for decision-making process, enabling experts to detect noticeable patterns and take advantage of them. One of the most common types of data is tabular data, which presents in almost every domains from economics, finance to healthcare, demography. Being organized in structured rows and columns, one can straightforwardly apply statistical methods, perform calculations and draw meaningful insights from this data. Moreover, many machine learning algorithms, especially those used in supervised learning tasks, are designed to work optimally on tabular data.

Since the beginning of the data-oriented era, missingness has always been of concern. Obviously, it is an annoying problem, potentially causing bias in the training process and leading to substandard results. Additionally, many analytical methods are either affected by these missing data, or even required to be performed on complete data, such as principal component analysis or singular value analysis. For tabular data, the missing issue arises when some cells in the table are left blank. This can stem from mistakes in the process of gathering, manipulating or analyzing data. In plenty of other cases, missing data can be attributed to the difficulties in collecting process (such as biospy data \cite{yoon2018gain}), or the collecting method of the machine \cite{troyanskaya2001missing}.

To alleviate the above detriments, people have tried several ways to eliminate the effects of missing data. At its simplest, one can completely remove the rows or the columns with missing. However, this approach not only wastes a great amount of valuable data, but also risks losing important infomation reflected in the discarded data. Another popular remedy is assigning values to missing cells using the mean, median or mode of the observed part of the corresponding feature, depending on its data type. This method is simple and fast, yet not optimal, since it uses a common value to all missing cells of a feature, without considering the unique characteristics of each sample, nor paying attention to the their correlations.

Besides those basic imputation methods, many studies have also tried to leverage machine learning models to cope with missing data. In the literature, they are often divided into two categories: discriminative models and generative models. Discriminative imputation methods usually based on supersived learning algorithms, where a model is trained with observed data, before being used to predict the unseen cells. By contrast, generative imputation models are trained to capture the characteristics of the dataset, thereby solving the task by synthesizing and generating new data. While discriminative models impute a distinct value for each missing cell, generative methods are more suitable for multiple imputation, thus able to capture the uncertainty of the assigned values.

With the thriving evolution of deep learning in the past decade, many generative models with different approaches had been proposed and become popular in both academia and industry. Some famous models are Gaussian Mixture Models (GMM), Hidden Markov Models (HMM), Variational Autoencoders (VAE) and Generative Adversarial Networks (GAN). These models are powerful tools for capturing the distribution of data, thus generating high-quality data for many purposes. Consequently, each of them has its corresponding versions for imputation task \cite{nazabal2020handling, mattei2019miwae, yoon2018gain}. Literature showed that they can effectively capture the distribution of observed data, thereby not only generating better assignments, but also improving the accuracy in the downstream tasks.

During the contemporary years, a new generative model called Neural Cellular Automata (NCA) \cite{mordvintsev2020growing} is drawing considerable attention. The original idea, Cellular Automata (CA), was initiated by John von Neumann in 1951 and widely studied in the later decades. CA is a computational system that is discrete and abstract, inspired by the morphogenesis process of organisms. Despite using simple and local rules, CA is able to model complex behaviors of dynamic systems in various scientific domains. Recently, \cite{gilpin2019cellular} showed that a deep convolutional neural network is fully capable of representing arbitrary two-dimensional CA. Moreover, \cite{mordvintsev2020growing} successfully trained a continuous NCA model to generate emoji patterns, which also offered many other impressive capabilities, such as the ability to persist results, as well to recover and regenerate after being damaged. This success has motivated more in-depth research on both the theoretical properties and pragmatic applications of NCA.

Knowing that NCA is a remarkable generative model, we question about an NCA-based model for imputing missing data. However, most of the related works merely concentrated on creating images or artifacts in two or three-dimensional spaces. This is reasonable, since NCA inherently synthesizes state information in a neighborhood to update cell's state, and working directly on low-dimensional space can make the results visualizable. Up to now, the existence of an imputation model based on NCA still remains an open question. Viewing data samples as pixels and their features as channels, solving the imputation task with an NCA-like model is very prospective.

Nevertheless, tabular data intrinsically does not exhibit local similarity like image data. That is, while neighboring pixels are often highly similar, data samples in a table generally do not possess this property. Thus, instead of synthesizing information in a neighborhood like NCA, the goal now becomes searching and aggregating information of data instances that are similar to the missing data instance. To do this, we need an architecture that is able to repetitively calculate similarity in the entire dataset and collect information from prominent samples. One may think that the Attention Mechanism in \cite{bahdanau2014neural, vaswani2017attention} sounds adequate to solve the problem, but we find that notable modifications are necessary to adapt the imputation task.

Overall, in this paper, we propose a novel imputation method inspired by the NCA model, which we call Neural Imputation Cellular Automata (NICA). Our model leverages the attention mechanism to iteratively find the ``neighborhood'' of each samples based on similarities and learn their ``growing rule'', thus regenerating values for the absent entries. Intuitively, after growing, the generated data would tend to cluster in separated groups of neighbors. It is also noteworthy that, unlike some imputation methods, NICA does not require the fully observed data in the training process. This is valuable, due to the fact that missing data might be an inherent problem in many practical circumstances. For these cases, it is infeasible to obtain the complete dataset, and imputation methods allowing missing input are apparently more beneficial.

Our contributions are: (i) We introduce NICA, a novel generative imputation model based on NCA architecture; (ii) We provide several experiments to compare the performance of our model with other imputing algorithms at different settings; (iii) We illustrate the clustering behavior of the data samples throughout the growing process; (iv) We report accuracy in downstream task when training with our imputed data versus other methods. To facilitate reproducibility, we provide the source code at: \href{https://github.com/TrungTin98/NICA}{\texttt{github.com/TrungTin98/NICA}}.

In the following Section 2, we review the noticeable imputation methods and NCA-related literature. Next, in Section 3, after stating the problem, we give a glance at the missingness mechanism and the attention mechanism. We then depict our method and necessary adjustments to the original attention in Section 4. Section 5 contains several experiment results and analysis, before our conclusion in the final section.

%% file: content/2.review.tex
\section{Literature review}

\subsection{Imputation methods}

In data pre-processing stage, handling missing data is a paramount step to warrant valid analysis and improve later performance. Besides the simple imputation methods such as assigning central tendency of corresponding attributes, machine learning models have been widely used since the beginning of this century. The KNNimpute method in \cite{troyanskaya2001missing} is based on k-nearest neighbors algorithm, which finds $k$ most similar samples using Euclidean distance and assigns values by their weighted mean. Later, in \cite{stekhoven2012missforest}, authors upgraded the above method with the random forest algorithm. The most prevalent methods until now are those using chained equations \cite{raghunathan2001multivariate, van2011mice}, where each feature is predicted consecutively by appropriate models trained with the remaining features. While \cite{raghunathan2001multivariate} sticked to regression models, \cite{van2011mice} allowed wider range of model, meticulously chosen depending on data type. Overall, these discriminative methods offer good performance, along with significant impact on relating tasks \cite{vo2024effects, pham2024correlation} and downstream problems \cite{hua2024impact}, yet only impute a sole value for each blank cell, neglecting the effect of variability.

Leveraging the unprecedented evolution of deep learning in the previous decade, advanced generative models are better than ever. Some renowned models include Gaussian Mixture Models (GMM), Hidden Markov Models (HMM), Variational Autoencoders (VAE) and Generative Adversarial Networks (GAN). Despite different approaches, they can effectively model patterns and relationships in data, hence generating high-quality data tailored for a variety of goals. However, an undeniable drawback of deep learning is that models are usually over-parameterized to achieve better performance, which may lead to unexplainable results and the risk of overfitting.

The missing data imputation problem can also be considered as a generative task, where models first learn the distribution of the observed data and then generate plausible values for the unobserved part of the table. Therefore, many studies have tried to use each of the aforementioned models to impute missing data. \cite{gondara2018mida} proposed MIDA, an overcomplete denoising auto-encoder that projected input data into a higher-dimensional space in order to capture its latent interactions and then recover missing information. In HI-VAE \cite{nazabal2020handling}, the authors adapted the VAE architecture \cite{kingma2013auto} to handle incomplete heterogeneous data. More specifically, their encoder tries to capture the joint distribution of the latent variable and the absent entries conditioned on the known data, while the decoder is a probabilistic model parametrized by a neural network whose input is the latent variable drawn from Gaussian mixture distribution. On the other hand, MIWAE \cite{mattei2019miwae}, as its predecessor IWAE \cite{burda2015importance}, focused on modifying the evidence lower bounds to ensure tightness and flexibility when latent variable does not follow Gaussian rule. Alternatively, \cite{yoon2018gain} trained a generative adversarial network where the generator tried to impute the missing data conditioned on the observed part and the discriminator attempted to determine whether each cell is actually missing or not.

Research has shown that the above generative imputation methods are able to assigned good values for missing data. Additionally, since they try to capture the distribution of data, one can draw multiple assignments for unobserved values after the training process. This is the concept of multiple imputations, which allows models to reduce the uncertainty of the imputed data. Even so, since they use deep intricate neural network to model the relationship of data, their architectures are usually extremely complex, going along with a huge amount of trainable parameters. To the end, simple model for imputation is always worth investigating.

\subsection{Neural Cellular Automata}

NCA is based on the Cellular Automata (CA) computational model, which was introduced by John von Neumann in \cite{von1951general} during his works on self-reproduction theory of biological evolution. His automaton was not the simplest, but the foremost discrete computational model proven to be a universal Turing machine. Later, a two-dimensional CA called Conway's Game of Life \cite{gardner1970mathematical} became the most famous automata model and the simplest universal computer ever. For more than seven decades, CA had drawn considerable attention in academia for both theoretical properties and practical applications. Especially, Stephen Wolfram gave many systematic studies on the one-dimensional CA model in \cite{wolfram2003new}. In short, CA attracts researchers because it is a powerful computational machine with the capability of simulating complex patterns. It is therefore applied in many fields of science comprising physics, bioinformatics, security and urban evolution. 

For simplicity, CA can be visualized as a grid of cells whose color changes over discrete time steps with a given rule. In essence, it is a dynamic system consisting of four components:
\begin{itemize}
    \item \textbf{Discrete states}: At each time step, each cell has its own independent, discrete and finite state, which is usually represented as a vector of numbers.
    \item \textbf{Discrete spaces}: All cells are located on a discrete lattice. A diversity of layout and number of dimension can be presupposed, and homogeneity is usually assumed;
    \item \textbf{Dynamics}: At each time step, the state of each cell changes according to predefined rules, which usually depend on the system's state at the previous step;
    \item \textbf{Local interactions}: The transformation of each cell only depends on the interactions of its neighboring cells.
\end{itemize}

In the last century, works on CA were dominated by cases with discrete states, discrete spaces and manageable rules due to the limitation of computational capability. Recently, with the thriving development of computers, researchers begin studying CA with continuous states and spaces, as well as more complicated dynamics. Specifically, instead of predefining a fixed rule for the evolution of CA, they can train a neural network to learn those rules. Indeed, \cite{gilpin2019cellular} showed that a deep convolutional neural network is fully capable of representing arbitrary two-dimensional CA. The authors also proposed an architecture consisting of a $3\times3$ convolutional layer before some $1\times1$ convolutional layers to learn CA rules. Later, in 2020, with some technical adjustments, \cite{mordvintsev2020growing} built a CA model with continuous state space to generate emojis from only one initial black cell, and called it NCA. Notably, this model also has many other impressive capabilities, such as the ability to maintain results for a long time and the ability to recover after being damaged.

This success has motivated further investigation into both the theory and applications of NCA. On the one hand, many articles continue to delve into its characteristics such as robustness \cite{randazzo2021adversarial}, isotropy \cite{mordvintsev2022growing} and steerability \cite{randazzo2023growing}. Additional capabilities of NCA like segmentation or self-classification have also been investigated \cite{sandler2020image, randazzo2020self}. On the other hand, NCA is applied as an effective generation model in plenty of problems in diverse fields such as synthesizing textures \cite{niklasson2021self}, growing 3D artefacts \cite{sudhakaran2021growing}, regenerating soft robots \cite{horibe2021regenerating}, creating game levels \cite{earle2022illuminating}. Moreover, some literatures have tried to design architechtures that combine NCA with other generative models including GAN \cite{otte2021generative}, VAE \cite{palm2022variational} or Vision Transformer \cite{tesfaldet2022attention}.

Though having achieved remarkable results, it is obvious that most of the current research merely focuses on generative tasks in two- or three-dimensional spaces, which related to the traditional CA. From our perspective, a powerful generative model like NCA may be potential for addressing the imputation problem. However, a study into this aspect is still lacking. One main obstacle of this is solving the question of locality, since tabular data do not offer similarity between adjacent instances. Here, our strategy is using suitable similarity kernel to determine the neighbors of each sample, thereby capturing the local interactions that promote the evolution of missing cells.

%% file: content/3.preliminary.tex
\section{Preliminary}

\subsection{Problem Formulation}

A fully observed tabular data can be represented as matrix $\mathbf{X} = (x_{ij}) \in \mathbb{R}^{n \times d}$, where each row $\mathbf{x}_{i \cdot} = (x_{i1}, \dots, x_{id})$ with $i=1,\dots,n$ is a sample and each column $\mathbf{x}_{\cdot j} = (x_{1j}, \dots, x_{nj})$ with $j=1,\dots,d$ is a feature. Define $\mathcal{R} = \mathbb{R} \cup \{\ast\}$ where $\ast$ denotes missing value, which is not an element of $\mathbb{R}$. Let $\tilde{\mathbf{X}} = (\tilde{x}_{ij}) \in \mathcal{R}^{n \times d}$ be the missing data described as
\begin{equation*}
    \tilde{x}_{ij} = \begin{cases}
        x_{ij}, & x_{ij} \text{ is observed},\\
        \ast,   & x_{ij} \text{ is unobserved}.
    \end{cases}
\end{equation*}
Now define matrix $\mathbf{M} = (m_{ij}) \in \{0;1\}^{n \times d}$ that satisfies
\begin{equation*}
    m_{ij} = \begin{cases}
        1, & x_{ij} \text{ is observed},\\
        0, & x_{ij} \text{ is unobserved}.
    \end{cases}
\end{equation*}
Then $\mathbf{M}$ is the mask matrix to indicate the location of missing cells in $\mathbf{X}$. Note that one can straightforwardly obtain $\mathbf{M}$ from $\tilde{\mathbf{X}}$.

\subsection{Missingness Mechanism}


In general, missing data is the part of the data that is unknown but would be valuable if it were known. Denote $\mathbf{X}^{obs}$ and $\mathbf{X}^{mis}$ respectively the observed part and the unobserved part of data. In general, the distribution of $\mathbf{M}$ is conditioned on $\mathbf{X}^{obs}$ and $\mathbf{X}^{mis}$. This relation is called the \emph{missing data model}
\begin{equation*}
    p \tron{\mathbf{M} \mid \mathbf{X}^{obs}, \mathbf{X}^{mis}, \boldsymbol{\theta}},
\end{equation*}
where $\boldsymbol{\theta}$ is the parameters of the model.

The most common way to classify missingness in the literature is based on the mechanism that leads to missingness. According to \cite{rubin1976inference, little2019statistical}, missing data can be classified into three types. First, data are \emph{missing completely at random (MCAR)} if $\mathbf{M}$ is independent of both $\mathbf{X}^{obs}$ and $\mathbf{X}^{mis}$. Then, missing data is only conditioned on model parameters:
\begin{equation*}
    p \tron{\mathbf{M} \mid \mathbf{X}^{obs}, \mathbf{X}^{mis}, \boldsymbol{\theta}} = p(\mathbf{M} \mid \boldsymbol{\theta}).
\end{equation*}

MCAR is an ideal assumption that is too strict. A looser assumption about missing data is when its distribution is only correlated with known data and uncorrelated with unknown data:
\begin{equation*}
    p \tron{\mathbf{M} \mid \mathbf{X}^{obs}, \mathbf{X}^{mis}, \boldsymbol{\theta}} = p(\mathbf{M} \mid \mathbf{X}^{obs}, \boldsymbol{\theta}).
\end{equation*}
In this case, the data is said to be \emph{missing at random (MAR)}. This is more general and more realistic than MCAR, and is usually the go-to choice for generative imputing models, since they try to capture the distribution of $\mathbf{M}$ based on $\mathbf{X}^{obs}$.

Finally, when the data are neither MCAR nor MAR, it is said to be \emph{missing not at random (MNAR)}. The reasons for missing data can now be diverse and one cannot shorten the missing data model $p \tron{\mathbf{M} \mid \mathbf{X}^{obs}, \mathbf{X}^{mis}, \boldsymbol{\theta}}$.

\subsection{Attention Mechanism}

The attention mechanism is a vital ingredient in sequence modeling, allowing the model to capture dependencies between input and output irrespective of their distance. Suppose that there are $n$ queries of length $d_k$, $m$ pairs of keys and values of length $d_k$ and $d_v$, respectively. The \emph{Attention} operator on the query matrix $\mathbf{Q} \in \mathbb{R}^{n \times d_k}$, key matrix $\mathbf{K} \in \mathbb{R}^{m \times d_k}$ and value matrix $\mathbf{V} \in \mathbb{R}^{m \times d_v}$ is
\begin{equation*}
    \operatorname{\texttt{Attention}} (\mathbf{Q}, \mathbf{K}, \mathbf{V}) = \operatorname{\texttt{softmax}} \tron{\frac{\mathbf{Q} \mathbf{K}^{\mathsf{T}}}{\sqrt{d_k}}} \mathbf{V} \in \mathbb{R}^{n \times d_v}.
\end{equation*}

In order to capture the joint information of the data samples with various representations, the \emph{Multi-head Attention} operator is usually used
\begin{align*}
    \operatorname{\operatorname{\texttt{MHA}}} (\mathbf{Q}, \mathbf{K}, \mathbf{V}) &= \vuong{ \mathbf{head}_1, \dots, \mathbf{head}_h } \mathbf{W}^O, \\
    \text{where}\quad\;\, \mathbf{head}_i &= \operatorname{\texttt{Attention}} \tron{\mathbf{Q} \mathbf{W}^Q_i, \mathbf{K} \mathbf{W}^K_i, \mathbf{V} \mathbf{W}^V_i},
\end{align*}
with $\mathbf{W}^Q_i \in \mathbb{R}^{d \times d_k}$, $\mathbf{W}^K_i \in \mathbb{R}^{d \times d_k}$, $\mathbf{W}^V_i \in \mathbb{R}^{d \times d_v}$ and $\mathbf{W}^O \in \mathbb{R}^{hd_v \times d}$, for $i=1,\dots,h$. The case $\mathbf{Q}$, $\mathbf{K}$ and $\mathbf{V}$ are identical, the above operator is called \emph{Self-Attention}, which makes each data sample attend on its similar samples. Self-Attention on matrix $\mathbf{X} \in \mathbb{R}^{n \times d}$ is defined by
\begin{equation}
    \operatorname{\texttt{SelfAttention}} (\mathbf{X}) = \operatorname{\operatorname{\texttt{MHA}}} (\mathbf{X}, \mathbf{X}, \mathbf{X}).
\end{equation}

%% file: content/4.methodology.tex
\section{Methodology}

\begin{figure}[!t]
    \centering
    
    \begin{subfigure}[b]{1\textwidth}
         \centering
         \includegraphics[width=\textwidth]{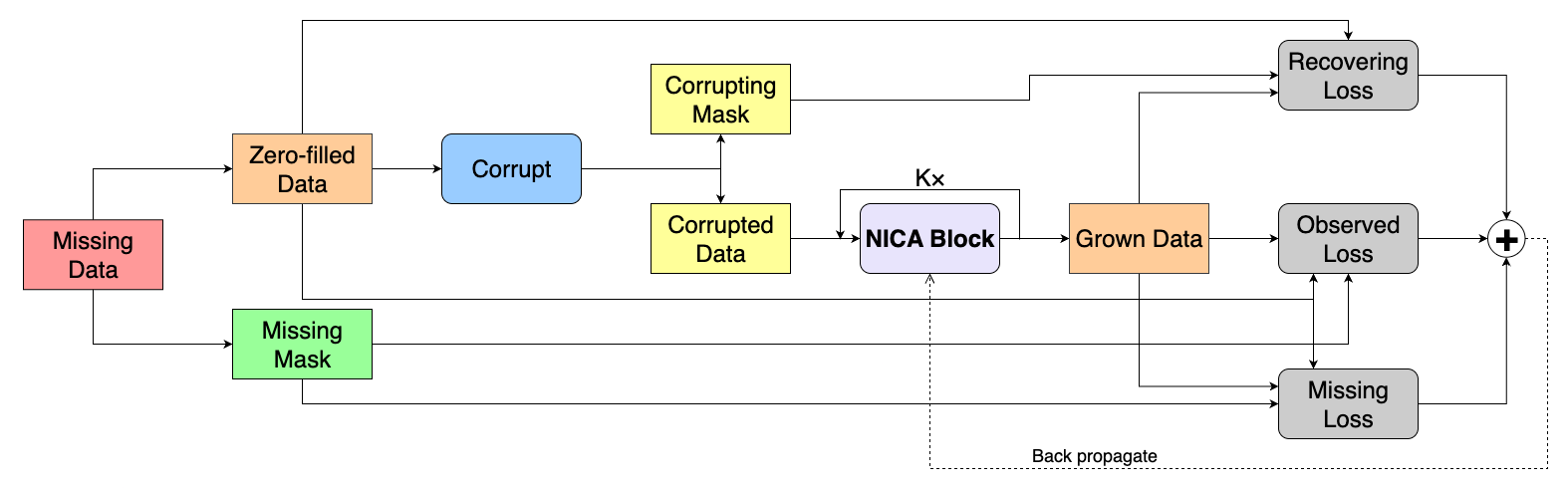}
         \caption{The processing flow in NICA}
         \label{NICA:flow}
    \end{subfigure}
    \begin{subfigure}[b][3cm]{0.8\textwidth}
         \centering
         \includegraphics[width=\textwidth]{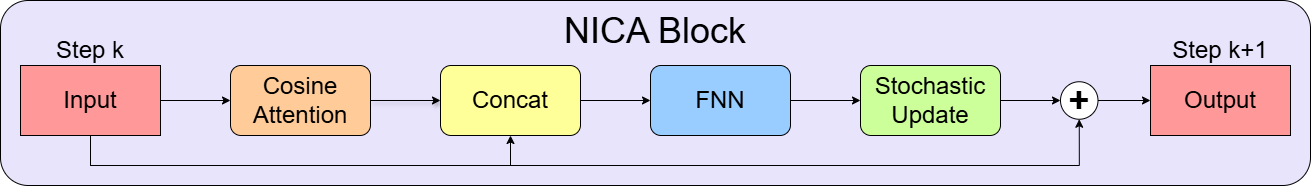}
         \caption{The architechture of NICA Block}
         \label{NICA:block}
    \end{subfigure}

    \caption{Our NICA model}
    \label{fig:NICA}
\end{figure}

In this section, first, we briefly summarize the main part of NCA architecture, which inspires our model. Then we describe our imputation model, which we call \emph{NICA - Neural Imputation Cellular Automata}. After that, some crucial modifications are presented and explained. An overall visualization of the proposed model is provided in Figure \ref{fig:NICA}.

\subsection{Model}

For NCA, input data is a grid that contains some one-valued cells in the center and zeros at the remaining cells, which is called seed. The architecture uses convolutional layers with predefined kernels, including fixed identity kernel and Sobel filters to model the cell state and the relations between neighboring cells. This information then goes through trainable dense layers, which are used to learn the dynamic of cells state. The process is repeated multiple times, which is called the growing process, and the number of growing steps is predetermined. The loss is then calculated by the difference between the output data and the target data, thereby performing back-propagation and optimizing the parameter set using gradient-based methods.

For the missing data problem, the imputation model needs to capture information from the entire dataset, before finding similar data points which play the role of ``neighbors''. Therefore, the convolutional layers in NCA are replaced by the self-attention layers. Besides, two additional losses that are critical in imputation task are also taken into consideration. The first one would ensure that the imputing values at the known positions are close to the observed data, while the second tried to pushed the imputing values at the unseen cells far away from the initial assignment. These losses not only guarantee good imputation, but also prevent our model from trivial solution.

More particularly, similar to other imputation methods, NICA receives as input a matrix $\tilde{\mathbf{X}}$, which contains blank cells. First, it creates a matrix $\mathbf{M}$ of the same size as $\tilde{\mathbf{X}}$ to mark the observed positions. Next, the data is normalized and pre-assigned with zeros to create a simple baseline version denoted by $\bar{\mathbf{X}}_0$. After that, the initial seed is constructed by randomly corrupting a certain proportion of the known cells of $\bar{\mathbf{X}}_0$. Now, we have a corrupted data $\mathbf{X}_c$ and its corresponding corrupting mask $\mathbf{M}_c$.

The training phase begins by passing $\mathbf{X}_c$ through self-attention layer where attention weights are calculated, so that NICA can automatically determine neighboring samples and output the interactions between them. These information is then concatenated with the current state, before being mapped to the space with the same dimension of $\mathbf{X}_c$ by learnable feedforward neural network to achieve the growing dynamic. This procedure is repeated for a preset times $K$. We sum up the whole process by
\begin{align*}
    \mathbf{X}_c^{(0)} &= \mathbf{X}_c, \\
    \Delta \mathbf{X}_c^{(k)} &= \operatorname{\texttt{FNN}} \tron{\operatorname{\texttt{stack}}\vuong{\operatorname{\texttt{SelfAttention}} \tron{\mathbf{X}_c^{(k)}}, \mathbf{X}_c^{(k)}}}, \\
    \mathbf{X}_c^{(k+1)} &= \mathbf{X}_c^{(k)} + \Delta \mathbf{X}_c^{(k)},
\end{align*}
for $k = 0,\dots,K-1$ and the $\operatorname{\texttt{FNN}}$ operation consists of linear layers with appropriate activation function and dropout layers. 

After the loop, grown data $\mathbf{X}_c^{(K)}$ is used to calculate losses. Our objectives are (i) recovering missing data from the corrupted data, (ii) generating values that are as close as observed data as possible, and (iii) staying away from trivial solution. For the first purpose, we calculate the \emph{recovering loss}, which is the difference between the grown data and our zero-imputed version at the corrupted cells
\begin{equation*}
    \mathcal{L}_{recovering} = \mathbb{E}_{\mathbf{M}_c} \vuong{\mathcal{L} \tron{\mathbf{X}_c^{(K)}, \bar{\mathbf{X}}_0}},
\end{equation*}
where $\mathcal{L} \tron{\mathbf{X}, \mathbf{Y}}$ is the squared difference between two matrices $\mathbf{X}$ and $\mathbf{Y}$. This loss is partly adopted from the objective in training NCA architecture. Then, a dominant loss in missing data imputation literature is the dissimilarity between generated data and known data, which we call \emph{observed loss}
\begin{equation*}
    \mathcal{L}_{observed} = \mathbb{E}_{\mathbf{M}} \vuong{\mathcal{L} \tron{\mathbf{X}_c^{(K)}, \bar{\mathbf{X}}_0}}.
\end{equation*}
The second goal is straightforwardly reached with this loss. Finally, an additional loss on unobserved cells is apllied in order to push the grown values at missing cells far away from their initialization. It penalizes these cells if the generated data at the posterior steps does not change significantly compared to its value at the first step. Our \emph{missing loss} is calculated by
\begin{equation*}
    \mathcal{L}_{missing} = - \mathbb{E}_{\mathbf{1-M}} \vuong{\mathcal{L} \tron{\mathbf{X}_c^{(K)}, \bar{\mathbf{X}}_0}}.
\end{equation*}

Hence, the overall loss function of NICA is
\begin{equation*}
    \mathcal{L}_{model} = \alpha_1 \mathcal{L}_{recovering} + \alpha_2 \mathcal{L}_{observed} + \mathcal{L}_{missing},
\end{equation*}
where $\alpha_1$ and $\alpha_2$ are the hyperparameters to adjust the contribution of $\mathcal{L}_{recover}$ and $\mathcal{L}_{observed}$ in comparison with $\mathcal{L}_{missing}$. When $\mathcal{L}_{model}$ has been optimized, the generated data is $\bar{\mathbf{X}}$, and the assigned data is calculated by
\begin{equation*}
    \hat{\mathbf{X}} = \mathbf{M} \odot \bar{\mathbf{X}}_0 + (\mathbf{1-M}) \odot \bar{\mathbf{X}},
\end{equation*}
where $\odot$ denotes element-wise multiplication. Eventually, $\hat{\mathbf{X}}$ is renormalized and the categorical features are rounded.

\subsection{Modifications of the Attention}

While existing NCA works on visual tasks can express their results in images and videos, NICA has to compete with other methods in terms of imputing error. Therefore, it may suffer higher potential of overfitting. If NICA learn the observed data more carefully than necessary, its generated values for the missing part might be inappropriate. Here, we present some considerable adjustments, including the choice of mapping and attention score function in attention, together with techniques to secure stabity and avoid overfitting. An overview of NICA is provided by the pseudocode in Algorithm \ref{alg:NICA}.

\input{table/algo1}

In the very first work on NCA, fixed Sobel kernels and identity kernel were used to model the state of cells and their local interactions. Recently, research has taken advantage of the Laplace filter to model the second-order interaction between cells \cite{pajouheshgar2024mesh}. From the attention-based perspective, high-order interaction can be attained by multihead self-attention \cite{song2019autoint}. Many works on tabular representation have also pointed out that multihead attention is useful. In contrast to this pattern, we find that single-head attention is sufficient for our case. Moreover, projecting data to query, key and value spaces is also unnecessary. This can be analogous to the fixed kernels in NCA, which only takes a few predefined types of interactive weights into account. Nevertheless, NICA does not strictly compel samples to depend on their adjacent rows, but allows them to find their own neighbors dynamically by recalculating similarity at every growing step.

Since NICA grows missing data consecutively multiple times before back-propagation, it may suffer the risk of exploding gradients, which results in unstable training and improper convergence. Possible ways to deal with this problem are normalizing the input of the attention and scaling the dot product, which are used in many other literature on attention. Though, to draw interactions from the product of attention weights and sample state, the value in attention should not be normalized. Hence, we choose cosine similary kernel, which can calculate pairwise scores and simultaneously normalize them. In particular, for query $\mathbf{q}_i \in \mathbb{R}^d$ and key $\mathbf{k}_j \in \mathbb{R}^d$, the attention score is
\begin{equation*}
    c(\mathbf{q}_i, \mathbf{k}_j) =\frac{\mathbf{q}_i^T \mathbf{k}_j}{\chuan{\mathbf{q}_i}_2 \chuan{\mathbf{k}_j}_2},
\end{equation*}
where $\chuan{\cdot}_2$ is Euclidean norm. Then, for $\mathbf{Q} \in \mathbb{R}^{n \times d_k}$, $\mathbf{K} \in \mathbb{R}^{m \times d_k}$ and $\mathbf{V} \in \mathbb{R}^{m \times d_v}$, our \emph{Cosine Attention} operator is
\begin{equation*}
    \operatorname{\texttt{CosineAttention}} (\mathbf{Q}, \mathbf{K}, \mathbf{V}) = \operatorname{\texttt{softmax}} \tron{{\sqrt{d_k}} \operatorname{\texttt{cosine}} \tron{\mathbf{Q}, \mathbf{K}}} \mathbf{V}.
\end{equation*}
where the matrix $\operatorname{\texttt{cosine}} \tron{\mathbf{Q}, \mathbf{K}} = \vuong{c(\mathbf{q}_i, \mathbf{k}_j)}$ contains cosine similarity between $\mathbf{q}_i \in \mathbf{Q}$ and $\mathbf{k}_j \in \mathbf{K}$, for $i = 1, \dots, n$ and $j = 1, \dots, n$. Here, the scaling factor $\sqrt{d_k}$ is inspired by the scaled dot product, noting that if the components of $\mathbf{q}_i$ and $\mathbf{k}_j$ are independent random variables with mean $0$ and variance $1$, their cosine similarity has mean $0$ and variance $d_k^{-1}$.

While in conventional CA, every cell grows synchronously over time, the first NCA models introduced the asynchronous growth. That is, at each step, a particular proportion of cells is kept unchanged by an updating mask. NICA also mimics this action, using the $\operatorname{\texttt{Dropout1d}}$ layer, which randomly neglects some rows at each step. However, for imputing task, this also induces undesirable fluctuation in results, which reduces performance. Thus, proposed model would not drop a high rate, so that the growing process can ensure consistency. 

Finally, as another way to maintain stability is training with multiple versions of corrupting data. Specifically, instead of merely corrupting as described above, the model creates many corrupting variants from the pre-assigned data, where each of them is corrupted independently from the others, denoted by $\vuong{\mathbf{X}_{c1}, \dots, \mathbf{X}_{cv}}$. These variants are then shuffled  and cheated as initial seeds for training. After that, the recovering loss would be calculated on their corresponding version of corrupting masks. Viewing these versions as a 3-dimensional tensor, NICA can be trained on all of them at once. Furthermore, training process can even be accelerated when using graphic cards. In this way, NICA not only secures consistent result and avoids overfitting, but also gives good imputation for new missing data, which is crucial when it comes to generative model.



%% file: table/algo1.tex
\begin{algorithm}[!t]
    \caption{NICA pseudo-code}\label{alg:NICA}
    \begin{algorithmic}
    
    \State \#\# Preprocessing
    \State Create missing mask $\mathbf{M}$ to mark the missing cells in imput $\tilde{\mathbf{X}}$.
    \State Normalize $\tilde{\mathbf{X}}$ and pre-assign $0$.
    \State Create $v$ versions of corrupted data $\mathbf{X}_c = \vuong{\mathbf{X}_{c1}, \dots, \mathbf{X}_{cv}}$ and corrupting masks $\mathbf{M}_c = \vuong{\mathbf{M}_{c1}, \dots, \mathbf{M}_{cv}}$.
    \State 
    \State \#\# Training
    \For{$i \gets 1, n\_iteration$}
    \State $\mathbf{X}_c^{(0)} \gets \operatorname{\texttt{shuffle}} (\mathbf{X}_c)$
    \For{$k \gets 1, K$}
        \State $neighboring\_interaction\_X \gets \operatorname{\texttt{CosineAttention}} \tron{\mathbf{X}_c^{(k)}, \mathbf{X}_c^{(k)}, \mathbf{X}_c^{(k)}}$
        \State $all\_information\_X \gets \operatorname{\texttt{stack}} \tron{neighboring\_interaction\_X, \mathbf{X}_c^{(k)}}$
        \State $dynamic\_X \gets \operatorname{\texttt{FNN}} (all\_information\_X)$
        \State $\mathbf{X}_c^{(k+1)} \gets \mathbf{X}_c^{(k)} + \operatorname{\texttt{Dropout1d}} (dynamic\_X)$
        \EndFor
        \State Calculate loss function $\mathcal{L}_{model}$
        \State Update parameters in $\operatorname{\texttt{MLP}}$ using gradient-based method
    \EndFor
    \State 
    \State \#\# Imputing
    \State Generate values for all cells $\bar{\mathbf{X}}$.
    \State Calculate assigned data $\hat{\mathbf{X}} \gets \mathbf{M} \odot \tilde{\mathbf{X}} + (\mathbf{1-M}) \odot \bar{\mathbf{X}}$.
    \State Renormalize and round the categorical features.
    
    \end{algorithmic}
\end{algorithm}

%% file: content/5.experiments.tex
\section{Experiments}

Here we conduct several experiments to evaluate the performance of NICA in comparison with some of the most popular imputation methods. The experiments are designed to show that NICA offers lower imputation error and higher consistency in results compared to state-of-the-art imputation models. Moreover, when using imputed data for downstream tasks, output of our method is on par with the alternatives. The experiments is carried out at various missing rate, and results indicates the domination of NICA.  

\subsection{Experimental Design}

In our experiments, missing data is standardized and $20\%$ of the observed part is randomly corrupted for $v=8$ times. We train each data $1000$ iterations with the batch size of $1024$, and the number of growing steps is $K=10$. Since the $\operatorname{\texttt{CosineAttention}}$ outputs a tensor with the same shape of its input, output of $\operatorname{\texttt{stack}}$ is double the original dimension. The $\operatorname{\texttt{FNN}}$ is a fully-connected feedforward network comprising $2$ linear layers with ReLU activation and a dropout rate of $0.5$. The first linear layer maps the state information to the hidden space, whose dimension if $5$ times higher than that of the original, then the later one projects its back to original space. At each growing step, $10\%$ of samples is skipped from updating. When calculating model loss, we choose $\alpha_1 = \alpha_2 = 10$ to emphasize the importance of $\mathcal{L}_{recovering}$ and $\mathcal{L}_{observed}$ in imputation task. We train our model by optimizing $\mathcal{L}_{model}$ using Adam algorithm \cite{kingma2014adam} with learning rate $0.001$.

\subsubsection{Benchmark datasets and preprocessing}

Datasets in our experiments are well-known datasets from UCI \cite{uci} and Kaggle, which are widely used in missing data imputation literature. A brief description for them is given in Table \ref{data:info}, arranged in ascending order of number of data samples. We consider datasets with fewer than 1000 samples as small, those ranging from 1000 to 5000 samples as medium, and anything above as large. All datasets are fully observed, and we simply preprocess them by encoding the categorical variables to numerical type. Unless otherwise stated, $40\%$ of data is removed randomly following MCAR. NICA is able to handle missing data without any prior analysis.

\input{table/tab1}

\subsubsection{Baselines}

We compare our model with the following methods:

\begin{itemize}
    \item \textbf{Mean Imputation:} For baseline, we assign mean value of each column to its blank cells. This is the simplest and fastest way to impute missing data that is widely use in practice.
    \item \textbf{KNNimputer:} We use a famous imputing algorithm proposed in \cite{troyanskaya2001missing}, which is based on the $k$ nearest neighbors algorithm.
    \item \textbf{MICE:} After being proposed in \cite{van2011mice}, this method has become baseline in numerous research. It iteratively uses regression models to predict missing values of each feature conditioned on the counterparts.

    The above discriminative methods are applied via scikit-learn library \cite{scikit-learn}. All the configurations are set as default.
    \item \textbf{GAIN:} For generative method, we use the generative imputation adversarial networks in \cite{yoon2018gain}. This method is extremely popular in missing data literature. Configuration is as same as that was used in the original paper.
\end{itemize}


\subsection{Results and discussion}

\subsubsection{Imputation performance}

In this experiments, from the fully observed data, we create missingness by randomly cover $40\%$ of cells. Models are trained with this missing data, before being used to generate imputation. We then calculate the root mean square errors (RMSE) between the imputed data and the original data at the missing positions. Lower RMSE means better imputation, since it indicates that the imputed values and original values are similar. We do this 10 times for each dataset and report averages and standard deviations of the RMSEs in Table \ref{tab:rmse}.

\input{table/tab2}

The provided results clearly demonstrate the superior performance of NICA in comparison with the counterparts in most of datasets. It is evident that our method achieves the lowest RMSE in $14$ out of the $15$ datasets, indicating its robustness in handling missing values. Notably, in datasets such as Wine, Spambase, and California, NICA significantly outperforms the other methods, achieving $11-14\%$ lower error rates than the runner-ups. This improvement suggests that NICA is more effective at capturing the underlying structure of missing data, leading to better imputations. The lower standard deviations in many cases also indicate that NICA is more stable and consistent in its assignments compared to other techniques.

One of the key observations from the table is that traditional imputation methods like Mean and KNNimputer generally perform worse than more advanced methods such as MICE, GAIN, and NICA. While MICE and GAIN show improvements over simpler methods, they still fall short of NICA in most datasets. In particular, figures show that NICA consistently outperforms GAIN across all datasets, with improvements ranging from around $10\%$ to over $34\%$. Even in cases where NICA is not the absolute best, such as the Yacht dataset, its performance remains competitive and far superior to the trounced methods.

\begin{figure*}[t]
    \centering
    
    \includegraphics[scale=0.42]{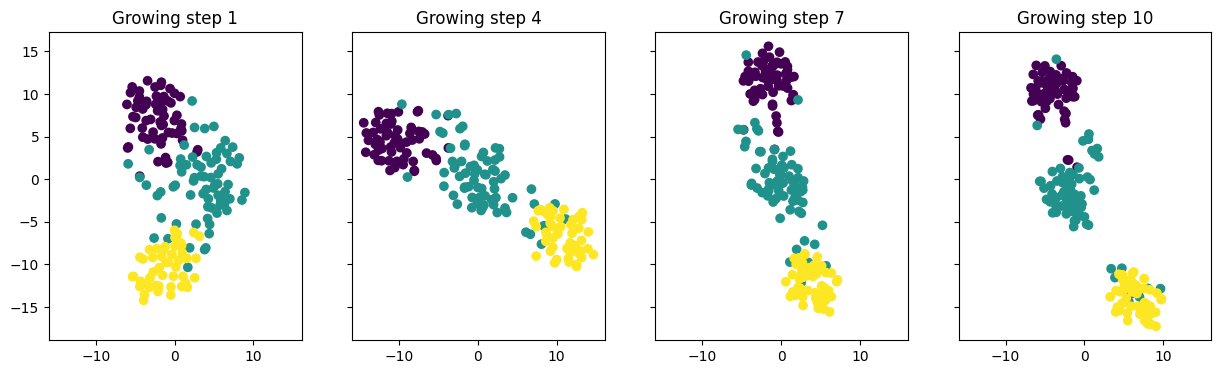}

    \caption{Data points form clusters through the growing process.}
    \label{fig:wine}
\end{figure*}

Collectively, the results position NICA as a state-of-the-art imputation technique, offering a balance of accuracy, consistency, and adaptability across datasets with varying characteristics. This evidence strongly supports the adoption of NICA as a preferred imputation method, especially in scenarios where minimizing reconstruction error is critical for predictive modeling. Beside that, in the growing process, data points have a tendency to find their own neighbors and thus gather into clusters. We illustrate this phenomenon in Figure \ref{fig:wine}, where the \emph{Wine} dataset is visualized using t-SNE \cite{van2008visualizing}.

\subsubsection{Post-imputing prediction performance}

Another popular experiment in imputation literature is post-imputing prediction, where imputed data is used to train a basic model, whose accuracy depicts the performance of the imputing methods. Here we use the imputed data from the above experiment to train regression models, depended on their target type. More particularly, we choose linear regression for data with continuous target and logistic regression for the rest. To ensure convergence, we simply normalize the imputed data before training. Linear model is then evaluated by R-squared, while the alternative is evaluated by the Area Under the Receiver Operating Characteristic Curve (AUROC) for binary cases and the extended AUROC One-vs-Rest (AUROC OvR) for multiclass cases. We do 5 cross validations on each imputed data, and report the average result in Table \ref{tab:accuracy}.

\input{table/tab3}

The results from Table \ref{tab:accuracy} highlight the effectiveness of the proposed method in enhancing downstream prediction performance across multiple datasets. NICA achieves the highest prediction score in $11$ out of $15$ datasets, demonstrating its ability to provide high-quality imputations that preserve the underlying data structure. Although NICA may not always be the top-performing method, it consistently maintains a competitive edge and significantly exceeds the performance of other approaches. The steady top-ranking performance of NICA suggests that it effectively captures missing data patterns, leading to more accurate predictions in regression models. Another notable observation is that NICA not only improves prediction performance but also provides stable results across datasets, as seen in its lower standard deviation compared to other methods in many cases. This stability is important in real-world applications where robustness and consistency are critical.

The results also depict the advantages of NICA over other methods regarding its scalability and versatility. Albeit our method sometimes takes second place in small datasets, it distinctly dominates the others in all $6$ medium and $2$ large datasets. 
For some small datasets such as \emph{Breast} and \emph{Blood}, MICE performs well, but it does not consistently outperform NICA. This suggests that the imputed values of MICE may induce linear dependencies, which may favor the chosen regression model. Therefore, it may not be as effective as NICA in handling more intricate missing data structures, as proven by the robustness of NICA across different types of data in the remaining datasets.

\subsubsection{Performance in different missing rates}
\begin{figure*}[!t]
    \centering
    
    \includegraphics[scale=0.27]{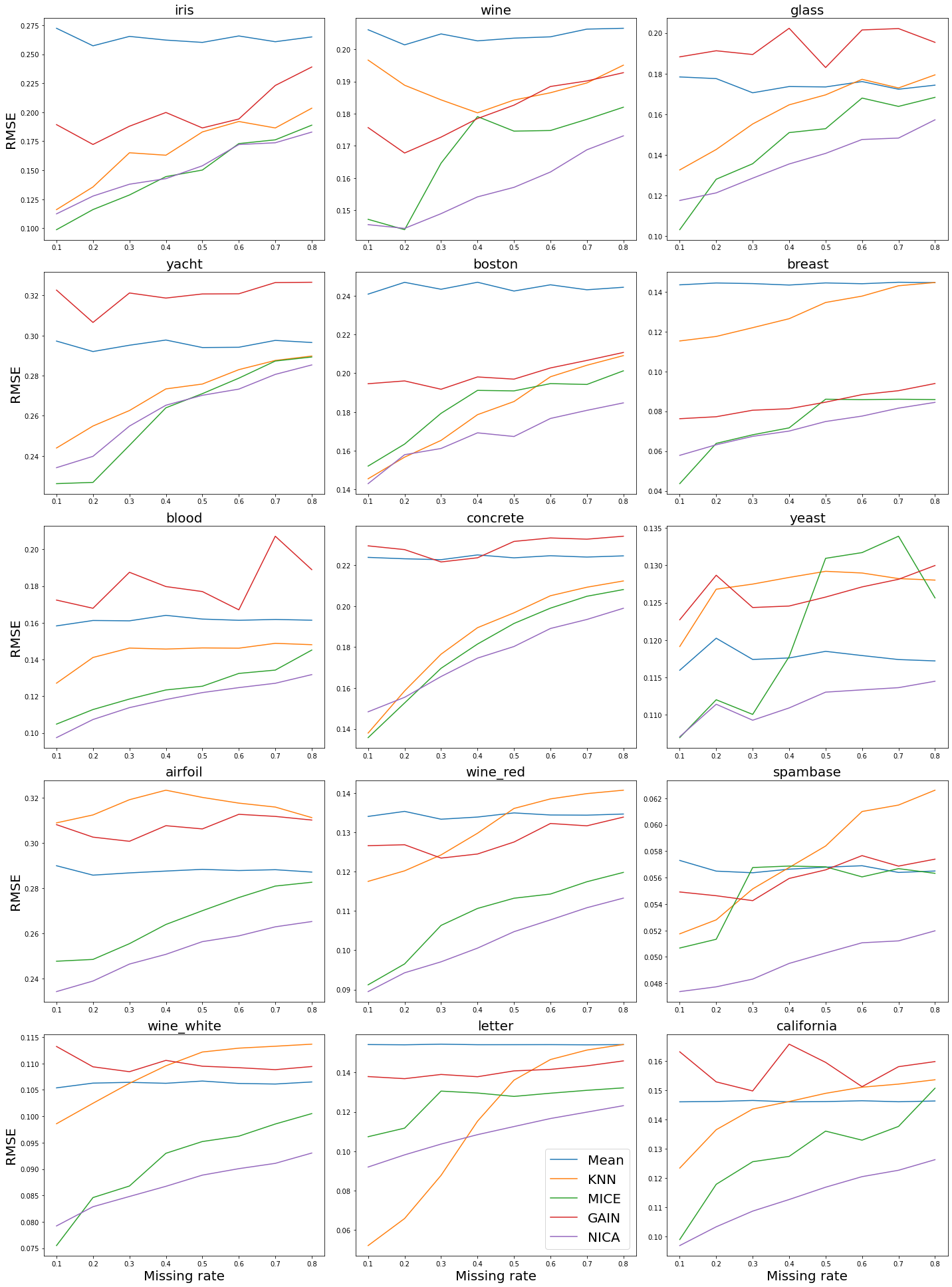}

    \caption{Average RMSE for various missing rates}
    \label{fig:rmse}
\end{figure*}

Here we repeat the above experiments with different missing rates ranging from $0.1$ to $0.8$. Figure \ref{fig:rmse} illustrates the RMSE performance of all methods across all $15$ datasets. A clear pattern emerges as NICA consistently achieves the lowest error across most datasets, demonstrating its superior capability to accurately estimate missing values. Notably, though some other methods perform reasonably well at low missing rates, they rapidly degenerate as missing rates escalate to over $30\%$. Meanwhile, NICA maintains relatively stable performance, underscoring its superiority in scenarios where severe missingness disrupts local or global patterns, ensuring solid imputation even under significant data loss.

\begin{figure*}[!t]
    \centering
    
    \includegraphics[scale=0.27]{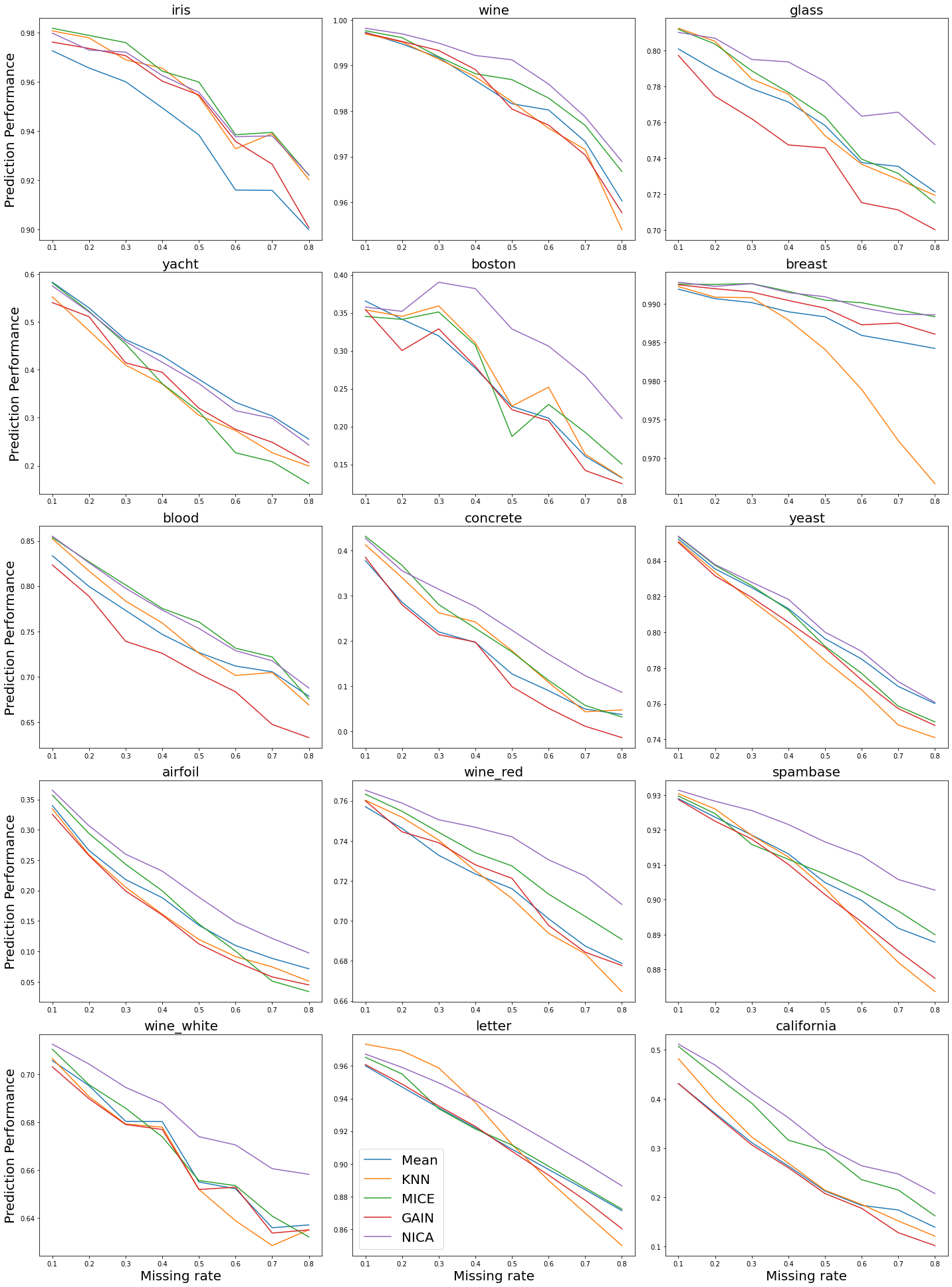}

    \caption{Average predicting performance for various missing rates}
    \label{fig:acc}
\end{figure*}

Figure \ref{fig:acc} presents the prediction performance of downstream models trained on data imputed by benchmark methods. Here, we observe a general decline in predictive accuracy as the missing rate increases, which is expected due to the loss of information. However, NICA mostly outperforms the other imputation techniques, especially at higher missing rates. This observation suggests that its imputed values preserve more meaningful relationships within the data, enabling better generalization in regression tasks. Furthermore, NICA also demonstrates significant improvement when dealing with medium and large datasets, as seen in the bottom two rows of Figure \ref{fig:acc}, where it consistently maintains higher prediction performance compared to other methods, particularly at higher missing rates, highlighting its robustness in handling large-scale and complex data distributions. This finding reaffirms the adaptability of NICA, making it an exceptionally valuable method in real-world applications where data incompleteness is pervasive.

%% file: table/tab1.tex
\begin{table*}[!t]
    \centering
    \caption{Dataset description}     
    \label{data:info}
    \begin{tabular}{|l||c|c|c|c|c|}
        \hline
        Dataset & \# Samples & \# Features & \# Continous & \# Categorical & \# Discrete \\
        \hline
        Iris & $150$ & $4$ & $4$ & $0$ & $0$ \\
        
        Wine & $178$ & $13$ & $11$ & $1$ & $1$ \\



        Glass & $214$ & $9$ & $9$ & $0$ & $0$ \\

        Yacht & $308$ & $6$ & $6$ & $0$ & $0$ \\

        Boston & $506$ & $13$ & $11$ & $1$ & $1$ \\

        Breast & $569$ & $30$ & $0$ & $0$ & $0$ \\

        Blood & $748$ & $4$ & $4$ & $0$ & $0$ \\

        Concrete & $1030$ & $8$ & $6$ & $0$ & $2$ \\

        Yeast & $1484$ & $8$ & $8$ & $0$ & $0$ \\

        Airfoil & $1503$ & $5$ & $3$ & $1$ & $1$ \\

        Wine Red & $1599$ & $11$ & $11$ & $0$ & $0$ \\
        
        Spambase & $4601$ & $57$ & $57$ & $0$ & $0$ \\
        
        Wine White & $4898$ & $11$ & $11$ & $0$ & $0$ \\

        Letter & $20000$ & $16$ & $0$ & $0$ & $16$ \\

        California & $20640$ & $8$ & $4$ & $1$ & $3$ \\


        \hline
    \end{tabular} 
\end{table*}

%% file: table/tab2.tex
\begin{table*}[t]
    \centering
    \caption{Imputation performance in terms of RMSE (Average ± Std of RMSE)} 
    \small
    \label{tab:rmse}
    \begin{tabular}{|l||c|c|c|c|c|}
        \hline
        Dataset & \textbf{Mean} & \textbf{KNN} & \textbf{MICE} & \textbf{GAIN} & \textbf{NICA} \\ \hline
        Iris & $0.262 \pm 0.005$ & $0.163 \pm 0.015$ & $0.145 \pm 0.008$ & $0.200 \pm 0.025$ & $\mathbf{0.143 \pm 0.011}$ \\ \hline
        Wine & $0.203 \pm 0.003$ & $0.180 \pm 0.004$ & $0.179 \pm 0.006$ & $0.178 \pm 0.007$ & $\mathbf{0.154 \pm 0.004}$ \\ \hline
        Glass & $0.174 \pm 0.010$ & $0.166 \pm 0.011$ & $0.151 \pm 0.014$ & $0.202 \pm 0.024$ & $\mathbf{0.135 \pm 0.014}$ \\ \hline
        Yacht & $0.297 \pm 0.006$ & $0.273 \pm 0.010$ & $\mathbf{0.264 \pm 0.012}$ & $0.319 \pm 0.016$ & $0.265 \pm 0.006$ \\ \hline
        Boston & $0.247 \pm 0.004$ & $0.179 \pm 0.003$ & $0.191 \pm 0.011$ & $0.198 \pm 0.008$ & $\mathbf{0.169 \pm 0.007}$ \\ \hline
        Breast & $0.143 \pm 0.002$ & $0.126 \pm 0.002$ & $0.072 \pm 0.005$ & $0.081 \pm 0.002$ & $\mathbf{0.070 \pm 0.002}$ \\ \hline
        Blood & $0.164 \pm 0.003$ & $0.146 \pm 0.008$ & $0.123 \pm 0.005$ & $0.180 \pm 0.023$ & $\mathbf{0.118 \pm 0.005}$ \\ \hline
        Concrete & $0.225 \pm 0.002$ & $0.190 \pm 0.002$ & $0.182 \pm 0.003$ & $0.224 \pm 0.006$ & $\mathbf{0.175 \pm 0.003}$ \\ \hline
        Yeast & $0.118 \pm 0.003$ & $0.128 \pm 0.004$ & $0.118 \pm 0.008$ & $0.125 \pm 0.007$ & $\mathbf{0.111 \pm 0.004}$ \\ \hline
        Airfoil & $0.288 \pm 0.002$ & $0.323 \pm 0.005$ & $0.264 \pm 0.003$ & $0.308 \pm 0.022$ & $\mathbf{0.251 \pm 0.003}$ \\ \hline
        Wine Red & $0.134 \pm 0.001$ & $0.130 \pm 0.002$ & $0.111 \pm 0.002$ & $0.124 \pm 0.006$ & $\mathbf{0.101 \pm 0.002}$ \\ \hline
        Spambase & $0.106 \pm 0.001$ & $0.110 \pm 0.001$ & $0.093 \pm 0.002$ & $0.111 \pm 0.004$ & $\mathbf{0.087 \pm 0.001}$ \\ \hline
        Wine White & $0.057 \pm 0.001$ & $0.057 \pm 0.001$ & $0.057 \pm 0.001$ & $0.056 \pm 0.001$ & $\mathbf{0.050 \pm 0.001}$ \\ \hline
        Letter & $0.154 \pm 0.000$ & $0.115 \pm 0.001$ & $0.130 \pm 0.002$ & $0.138 \pm 0.003$ & $\mathbf{0.108 \pm 0.001}$ \\ \hline
        California & $0.146 \pm 0.000$ & $0.146 \pm 0.001$ & $0.127 \pm 0.004$ & $0.166 \pm 0.015$ & $\mathbf{0.113 \pm 0.001}$ \\ \hline
    \end{tabular}
\end{table*}

%% file: table/tab3.tex
\begin{table*}[!t]
    \centering
    \caption{Prediction performance comparison} 
    \small
    \label{tab:accuracy}
    \begin{tabular}{|l||c|c|c|c|c|}
        \hline
        Dataset & \textbf{Mean} & \textbf{KNN} & \textbf{MICE} & \textbf{GAIN} & \textbf{NICA} \\ \hline
        Iris & $0.949 \pm 0.009$ & $\mathbf{0.966 \pm 0.013}$ & $0.964 \pm 0.007$ & $0.960 \pm 0.007$ & $0.963 \pm 0.007$ \\ \hline
        Wine & $0.987 \pm 0.004$ & $0.988 \pm 0.006$ & $0.988 \pm 0.005$ & $0.989 \pm 0.005$ & $\mathbf{0.992 \pm 0.003}$ \\ \hline
        Glass & $0.771 \pm 0.018$ & $0.776 \pm 0.018$ & $0.777 \pm 0.016$ & $0.748 \pm 0.028$ & $\mathbf{0.794 \pm 0.011}$ \\ \hline
        Yacht & $\mathbf{0.429 \pm 0.040}$ & $0.370 \pm 0.046$ & $0.371 \pm 0.090$ & $0.395 \pm 0.045$ & $0.416 \pm 0.039$ \\ \hline
        Boston & $0.277 \pm 0.050$ & $0.310 \pm 0.048$ & $0.307 \pm 0.078$ & $0.279 \pm 0.076$ & $\mathbf{0.382 \pm 0.038}$ \\ \hline
        Breast & $0.989 \pm 0.002$ & $0.988 \pm 0.002$ & $\mathbf{0.992 \pm 0.001}$ & $0.990 \pm 0.001$ & $0.991 \pm 0.001$ \\ \hline
        Blood & $0.747 \pm 0.010$ & $0.759 \pm 0.014$ & $\mathbf{0.775 \pm 0.009}$ & $0.726 \pm 0.036$ & $0.774 \pm 0.011$ \\ \hline
        Concrete & $0.196 \pm 0.027$ & $0.241 \pm 0.018$ & $0.228 \pm 0.036$ & $0.197 \pm 0.023$ & $\mathbf{0.276 \pm 0.026}$ \\ \hline
        Yeast & $0.813 \pm 0.004$ & $0.802 \pm 0.006$ & $0.812 \pm 0.009$ & $0.806 \pm 0.007$ & $\mathbf{0.818 \pm 0.005}$ \\ \hline
        Airfoil & $0.188 \pm 0.024$ & $0.161 \pm 0.026$ & $0.200 \pm 0.028$ & $0.160 \pm 0.021$ & $\mathbf{0.232 \pm 0.020}$ \\ \hline
        Wine Red & $0.723 \pm 0.013$ & $0.725 \pm 0.017$ & $0.734 \pm 0.013$ & $0.728 \pm 0.012$ & $\mathbf{0.747 \pm 0.012}$ \\ \hline
        Spambase & $0.913 \pm 0.002$ & $0.912 \pm 0.002$ & $0.912 \pm 0.004$ & $0.910 \pm 0.002$ & $\mathbf{0.922 \pm 0.002}$ \\ \hline
        Wine White & $0.680 \pm 0.014$ & $0.678 \pm 0.020$ & $0.674 \pm 0.013$ & $0.677 \pm 0.014$ & $\mathbf{0.688 \pm 0.012}$ \\ \hline
        Letter & $0.922 \pm 0.001$ & $0.938 \pm 0.001$ & $0.921 \pm 0.002$ & $0.923 \pm 0.002$ & $\mathbf{0.939 \pm 0.001}$ \\ \hline
        California & $0.264 \pm 0.021$ & $0.270 \pm 0.010$ & $0.316 \pm 0.036$ & $0.260 \pm 0.012$ & $\mathbf{0.362 \pm 0.022}$ \\ \hline
    \end{tabular}
\end{table*}

%% file: content/6.conclusion.tex
\section{Conclusion}

In this paper, we propose NICA, a generative method to impute missing tabular data. The algorithm is insprired by the NCA model in terms of growing the data for multiple steps, and is based on the self-attention mechanism to aggregate information. We benchmark our model on various real-world datasets, in several experiments, with different missing rates. The result shows that NICA outperforms not only a famous generative model, but also the canonical discriminative models. However, our model certainly has some limitations related to the attention architechture. For instance, when working on huge datasets, the required time to calculate attention weights might increase rapidly. Besides, these weights might be flatten, making each sample be equally dependent on all instances. A potential approach to tackle this problem is the temperature hyperparameter, which is used in some attention-related literature. Future work could improve the structure and details in NICA to achieve better imputation. In addition, while our method basicly preprocesses the categorical variables by encoding, upcoming studies can take non-trivial approaches such as feature embeddings and Transformer layers into consideration, which potentially lead to noticeable amelioration.